\documentclass[manuscript,screen]{acmart}

\AtBeginDocument{%
  \providecommand\BibTeX{{%
    \normalfont B\kern-0.5em{\scshape i\kern-0.25em b}\kern-0.8em\TeX}}}

\setcopyright{acmcopyright}
\copyrightyear{2014}
\acmYear{2024}
\acmDOI{XXXXXXX.XXXXXXX}

\usepackage{multirow}
\begin{document}

\title[Multistage Fine-tuning Strategies for Automatic Speech Recognition in Low-resource Languages]{Multistage Fine-tuning Strategies for Automatic Speech Recognition in Low-resource Languages}

\author{Leena G Pillai}
\authornote{Both authors contributed equally to this research.}
\email{leena.g@duk.ac.in}
\orcid{0000-0003-1692-7647}
\author{Kavya Manohar}
\authornotemark[1]
\email{kavya.manohar@duk.ac.in}
\orcid{0000-0003-2402-5272}
\affiliation{%
  \institution{Digital University Kerala}
  \city{Thiruvananthapuram}
  \state{Kerala}
  \country{India}
}

\author{Basil K Raju}
\affiliation{%
  \institution{Digital University Kerala}
  \streetaddress{Thiruvananthapuram}
  \city{Kerala}
  \country{India}}
\email{basil.cs21@duk.ac.in}

\author{Elizabeth Sherly}
\affiliation{%
  \institution{Digital University Kerala}
  \streetaddress{Thiruvananthapuram}
  \city{Kerala}
  \country{India}}
\email{sherly@duk.ac.in}


\begin{abstract}

This paper presents a novel multistage fine-tuning strategy designed to enhance automatic speech recognition (ASR) performance in low-resource languages using OpenAI’s Whisper model. In this approach we aim to build ASR model for languages with limited digital resources by sequentially adapting the model across linguistically similar languages. We experimented this on the Malasar language, a Dravidian language spoken by approximately ten thousand people in the Western Ghats of South India. Malasar language faces critical challenges for technological intervention due to its lack of a native script and absence of digital or spoken data resources. Working in collaboration with Wycliffe India and Malasar community members, we created a spoken Malasar corpus paired with transcription in Tamil script, a closely related major language. In our approach to build ASR model for Malasar, we first build an intermediate Tamil ASR, leveraging higher data availability for Tamil annotated speech. This intermediate model is subsequently fine-tuned on Malasar data, allowing for more effective ASR adaptation despite limited resources. The multistage fine-tuning strategy demonstrated significant improvements over direct fine-tuning on Malasar data alone, achieving a word error rate (WER) of 51.9\%, which is 4.5\%  absolute reduction when compared to the direct fine-tuning method. Further a WER reduction to 47.3\% was achieved through punctuation removal in post-processing, which addresses formatting inconsistencies that impact evaluation. Our results underscore the effectiveness of sequential multistage fine-tuning combined with targeted post-processing as a scalable strategy for ASR system development in low-resource languages, especially where linguistic similarities can be leveraged to bridge gaps in training data.

\end{abstract}





\maketitle

\section{Introduction}

The Malasar language, spoken by the indigenous Malasar community in the Western Ghats region of southern India, remains stable despite its relatively small speaker base\footnote{\url{https://www.ethnologue.com/language/ymr/}}. According to the 2011 census of India, there are only 9626 Malasar community members, out of which 6431 are in the state of Tamil Nadu and remaining are in the state of Kerala \cite{chandramouli2013scheduled}. Although the language is not sustained by formal institutions, it continues to thrive in homes and communities where it is passed down to children as the norm. 
As an oral language without a native script, Malasar is vulnerable to loss as communities diminish or integrate with larger linguistic groups. Documenting and archiving these languages become more complicated without a standardized script. In terms of technology integration, creating systems like automatic speech recognition (ASR) is significantly more difficult. ASR systems typically depend on large, annotated corpora for training, which are hard to generate without a written form of the language. This also affects the development of natural language processing (NLP) tools, such as machine translation and text-to-speech systems, which rely on a text corpus for training and evaluation. In general, the lack of a script poses a challenge for linguistic research. 

\begin{figure}[!h]
\begin{center}
\includegraphics[width=0.4\textwidth]{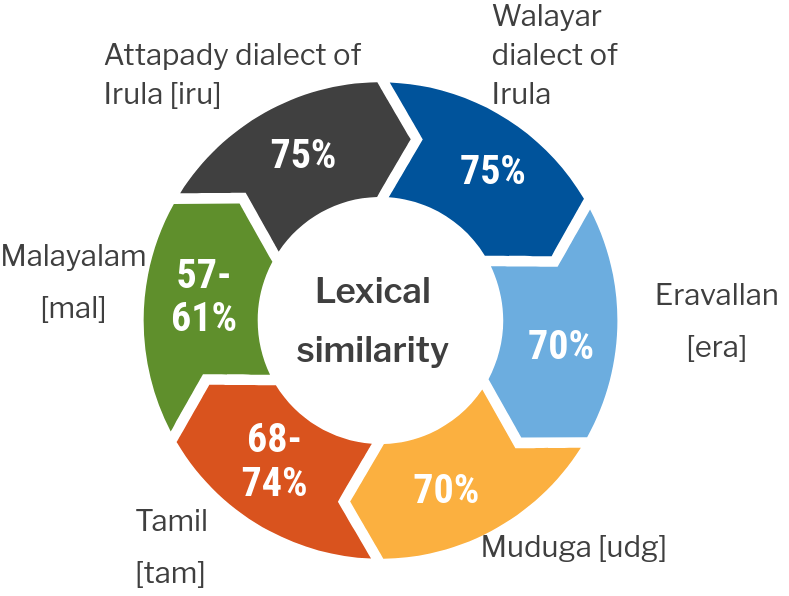} 
\caption{Lexical Similarity Of Malasar with other neighbouring languages}
\label{fig.lexi}
\end{center}
\end{figure}

To address the challenges of documenting unwritten languages, UNESCO published a set of guidelines in 2003
\cite{robinson2003writing}. These guidelines explore the processes involved in writing unwritten languages, thereby offering new opportunities for expression and learning to the world’s linguistic minorities and indigenous people. Malasar shares lexical similarities with Tamil, Malayalam, Muduga, Eravallan, and certain dialects of Irula \cite{bijumon2015tribes}. The lexical similarity of Malasar with other languages is illustrated in Fig. \ref{fig.lexi}. According to the guidelines \cite{robinson2003writing}, Tamil has been selected as the script for transcribing Malasar speech, facilitating a more consistent and comprehensible representation of the language.

The development of an ASR system for the Malasar language serves multiple critical purposes in language preservation and documentation efforts. By enabling accurate transcription of spoken Malasar into Tamil script, the system facilitates the creation of valuable linguistic records \cite{prud2021automatic} while supporting detailed analysis of the language's phonetic and grammatical features. Beyond documentation, an ASR system drives digital accessibility, fostering language revitalization through modern educational tools and enhanced community engagement. It also empowers governing bodies and strengthens the Malasar community's societal integration. In the context of low-resourced languages, such ASR development represents a crucial step toward preserving indigenous knowledge and cultural heritage. To address these specific linguistic and technical hurdles unique to building an ASR system for Malasar, our study adopts a multi-faceted approach:
\begin{enumerate}
    \item To build an initial corpus from scratch, we collaborated with Wycliff India, resulting in approximately four hours of transcribed Malasar audio data
    \item The lack of a native writing system for this oral-only language led us to adopt Tamil script for transcription due to its high lexical similarity, ensuring more accurate representation of Malasar speech
    \item To address data scarcity in training an ASR model, we implemented a novel sequential multistage fine-tuning strategy with Whisper, leveraging Tamil as an intermediate fine-tuning stage before adapting the model to Malasar
    \item To ensure accurate evaluation, we also introduced a punctuation filter, which significantly improved the reported WER by removing non-linguistic elements that impact the metric \cite{manohar2024lost}.
\end{enumerate}

In conlusion, the work described in this paper sets a foundation for further NLP applications in this under-resourced language.

\section{Literature Review}

Working with low resource languages presents a number of challenges, including data scarcity, tool scarcity, and expertise scarcity \cite{besacier2014automatic}. These challenges are generally addressed by using techniques such as data augmentation, transfer learning, and domain adaptation \cite{magueresse2020low,ranathunga2023neural}. The introduction of transformer models in ASR has indeed brought about a significant revolution in the field, especially for low-resource languages \cite{sherly2024asr}. ASR performance has improved drastically in recent years, mainly enabled by self-supervised learning based acoustic models such as Wav2vec2 \cite{schneider2019wav2vec, conneau2020unsupervised} and large-scale multilingual transformer models like Whisper \cite{radford2023robust}. 

Wav2vec2 is an encoder only transformer model. Using self supervised learning, this neural speech representation model   is trained with a lot of unlabelled data \cite{schneider2019wav2vec,baevski2020wav2vec}, in a process referred to as pre-training. For example the XSL-R model, that succeeded the Wav2vec2 model, was trained on 436K hours of publicly available unlabelled speech data from 128 languages \cite{conneau2020unsupervised}. The pretrained model could be adapted with comparably less amount of labelled data for a specific task like ASR, making it suitable for low-resource languages. The process of fine-tuning the pre-trained encoder model involves adding a linear classification layer on top of the transformer and training the entire model by minimising the connectionist temporal classification (CTC) loss function \cite{graves2006connectionist}. Whisper transformer model, on the other hand is a sequence-to-sequence model which consists of an encoder and a decoder linked via a cross-attention mechanism \cite{radford2023robust}. Unlike Wav2vec2 model, Whisper is trained on labelled speech data that amounts to  680K hours, of which 117K hours is non-English speech data in more than 90 world languages. 

There has been efforts in the past to fine-tune Wav2vec2 based encoder only models to transcribe endangerd languages of Nepal \cite{meelen-etal-2024-end} and field work corpus of Sino-Tibetan language family \cite{guillaume-etal-2022-fine}. Improvements in speech transcription accuracy was observed by incorporating additional text data for language modeling, when the amount of transcibed speech corpus is very low \cite{san-etal-2023-leveraging}. A comparison between encoder-only models and Whisper-like models for building ASR systems showed that pretrained models exposed to more data from a specific language family generally perform better on new, related languages within that family \cite{rouditchenko23_interspeech}. Attempts to fine-tune ASR systems for Dravidian languages from X-LSR \cite{kavya2023automatic} and Whisper \cite{chowdary2024transformer} models have shown effective improvements in accuracy. Whisper fine-tuning strategies in under resourced secenarios were studied in \cite{liu2024exploration} and has proved that fine-tuning with freezing the bottom layers has the strongest ability, while re-initializing the top layers is ineffective. This work has shown that adding bottleneck adapters and LoRA fine-tuning can significantly reduce computational and time costs, while sacrificing only a small amount of speech recognition ability \cite{liu2024exploration}. 


\section{Methodology}
The development of the Malasar ASR system involves several key phases, including dataset collection, preprocessing, model selection, fine-tuning, and evaluation. Given the challenges of limited data resources and the absence of a native script, we utilized the Tamil script as a practical representation for Malasar transcriptions. This choice facilitates data handling and leverages existing resources related to Tamil, a language with certain linguistic affinities to Malasar \cite{bijumon2015tribes}.

For the ASR task, we employed the Whisper model, a transformer-based architecture known for its multilingual capabilities. To evaluate the ASR performance on Malasar, we fine-tuned two versions of the Whisper model – Whisper Small (244M Parameters) and Whisper Medium (769M parameters)  – on the Malasar dataset across the following different configurations:
\begin{itemize}
    \item {\textbf{OpenAI Zeroshot (No Fine-tuning)}: In this baseline setup, we evaluated the original Whisper Small and Medium models on Malasar speech without any further fine-tuning. This zeroshot approach provided an initial benchmark for the models' ability to recognize Malasar without task-specific adaptation.}
    \item {\textbf{Direct Target Fine-tuning (DTF):} We fine-tuned Whisper Small and Whisper Medium directly on the Malasar dataset to adapt the models specifically for Malasar speech. This configuration aimed to enhance the models' performance by making them more sensitive to Malasar-specific phonetic and linguistic patterns.} 
    \item {\textbf{Multistage Target Fine-tuning (MTF) with Intermediate Tamil Pre-training:} To leverage linguistic similarities between Malasar and Tamil, we further evaluated both Whisper Small and Medium models by first performing intermediate fine-tuning on Tamil data, followed by target fine-tuning on Malasar. This multistage approach sought to improve ASR accuracy by utilizing transfer learning from a related language.}
\end{itemize}

Model performance was evaluated based on word error rate (WER) to measure the effectiveness of each configuration in adapting to the Malasar language. This methodology integrates strategic data processing, transfer learning and targeted fine-tuning to address the unique challenges posed by the Malasar language’s low-resource status and lack of a native script.

\subsection{Datacollection and Preprocessing}
Corpus plays a pivotal role in NLP by providing the foundational data required for training and evaluation of language models. Creating a suitable speech corpus involves the careful curation and transcription of audio recordings. 

\begin{table}[htpb]
  \caption{Malasar Speech Coupus}
  \label{table:corpus}
   \begin{tabular}{ll}
    \toprule
        Total Duration & 4:36:20 Hours \\
        Utterance & 965 \\
        Language & Malasar \\
        Transcription Script & Tamil\\
        Sample Rate & 16KHz \\
        Audio Length & Maximum 30 seconds \\
    \bottomrule
    \end{tabular}
\end{table}

The primary data for the Malasar corpus was collected by collaborating with Wycliffe India\footnote{\url{https://wycliffeindia.in/}}, a mission organization dedicated to spreading Bible preachings to all people in their native languages. The dataset consists of 5 hours of high-quality audio recordings, featuring speakers from the Malasar community. The recordings are gender-balanced, with speakers aged between 30 and 35. The audio content is derived from biblical texts.

The audio data was recorded in a controlled studio environment with minimal background noise and interference. This high-quality audio reduces the need for extensive noise removal. The presence of minor background noise, however, is acceptable for general ASR tasks, as it allows the model to be robust in real-world scenarios where perfect silence cannot always be guaranteed. Each audio file was furhter divided into segments of maximum 30 seconds, a duration chosen to balance between providing enough context for the model to learn from, while also ensuring that the segments are computationally manageable during training.  This segmentation is not only aligns with the transformer model's capacity but also prevents the model from overfitting on very short or very long utterances. 

The segmented audio was transcribed into Tamil script, in accordance with the guidelines for documenting unwritten languages \cite{robinson2003writing}. Malasar language has close lexical ties between neighbouring languages particularly Tamil and Irula. The Irula language lacks a native script and is typically written using the Tamil script. This existing practice served as a precedent for transcribing Malasar, which similarly lacks a written form. 

\begin{figure}[htpb]
    \begin{center}
        \includegraphics[width=0.4\textwidth]{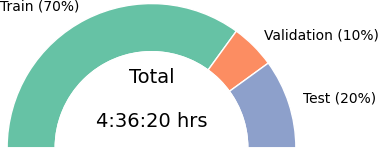}
        \caption{Audio data split used for train, validation and test.}
        \label{fig.audiosplit}
    \end{center}
\end{figure}

As discussed in Table \ref{table:corpus}, a total corpus of 4 hours, 36 minutes, and 20 seconds has been created for this work. Of this, 70\% (3 hours, 13 minutes, and 32 seconds) is used as training data for model fine-tuning, while 10\% (26 minutes and 43 seconds) has been allocated as the model validation set. Finally, the model was evaluated on a held-out test dataset, which constitutes 20\% of the total dataset, amounting to 54 minutes and 4 seconds. Figure \ref{fig.audiosplit} visually illustrates this audio data split.

\subsection{Direct Target Fine-tuning}

\begin{figure}[htpb]
    \begin{center}
        \includegraphics[width=0.9\textwidth]{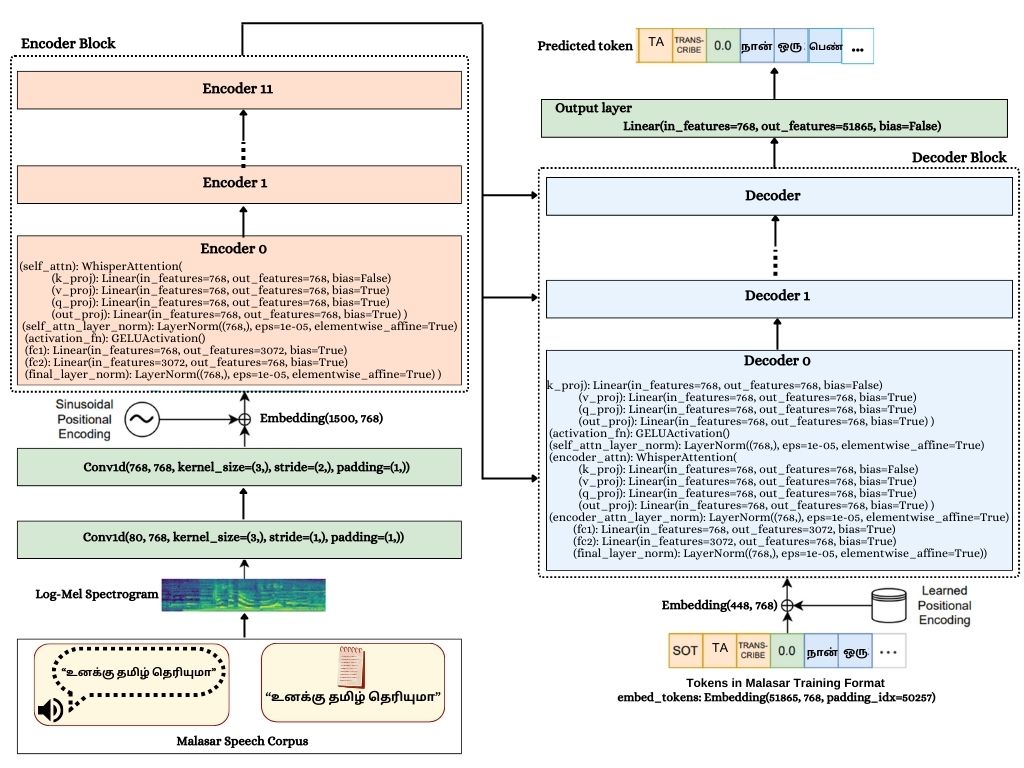}
        \caption{Schemantic reprersentation of DTF of Whisper architecture with Malasar}
        \label{fig.Mal_archi}
    \end{center}
    \vspace{-0.5cm}
\end{figure}

The OpenAI's Whisper architecture \cite{radford2023robust} is built upon a transformer-based encoder-decoder architecture, commonly used for sequence-to-sequence tasks, and is designed to process audio inputs and generate textual outputs like transcriptions or translations. The schematic representation of DTF of Whisper base model (small) with Malasar corpus is illustrated in Fig. \ref{fig.Mal_archi}.

The encoder processes the input audio and transforms it into a high-dimensional feature representation that is passed on to the decoder. In this base architecture, initial convolution takes the 80-dimensional mel spectrogram as input and applies a convolution operation to it, producing 768 feature maps. The kernel size is 3, meaning each output is influenced by a window of 3 time steps, and padding is used to ensure the output has the same length as the input. The second convolutional layer further processes the 768 feature maps and applies downsampling using a stride of 2. This effectively reduces the temporal resolution of the audio features by half, allowing the model to focus on more abstract patterns over time. The Whisper model uses positional embeddings, size of 1500, to encode the order of the input sequence. 

The encoder consists of 12 transformer layers, each comprising of self-attention mechanisms and feedforward networks, which is responsible for processing the input features and refining them into context-aware embeddings. To provide the model with the sense of sequence order, positional embeddings are used. Layer normalization is applied throughout the network to stabilize training and improve performance, ensuring that the encoder outputs well-formed representations that can be effectively utilized by the decoder.

The decoder generates the textual output from the encoded audio features. The decoder also consists of multiple components similar to the encoder but with additional attention layers to process both the input sequence and the output sequence being generated. The decoder also uses positional embeddings to track the order of the output tokens and passes its output through 12 transformer layers before projecting the final hidden states into the model’s vocabulary of 51,865 tokens. This large vocabulary makes the model adaptable to various languages and contexts, while the final output projection layer ensures accurate prediction of the next token in the sequence.

Whisper Medium version shares the same architecture as the small version, with key differences in scale. It uses a larger encoder and decoder with 24 transformer layers each, compared to 12 in the small model, enhancing its capacity to learn more complex patterns. The feature maps generated during convolutional processing increase to 1024 dimensions, allowing for a richer representation of the audio input. The transformer blocks within both the encoder and decoder have an increased number of parameters, enabling the model to capture long-range dependencies and perform more nuanced recognition tasks. 

The training configuration for fine-tuning the Whisper model on the Malasar ASR task is designed for optimal performance on low-resource speech data.
The training uses batch size of 32, with gradient accumulation set to 1, ensuring the model updates after each batch. The learning rate is set to $1e^{-5}$, a low value appropriate for fine-tuning, with 500 warmup steps to stabilize the training process. The total number of training steps is set at 2000. Choosing WER as the primary evaluation metric, the model is evaluated every 250 steps and the best-performing model (lowest WER) is saved at the end of training. 

\subsection{Multistage Target Fine-tuning}
\begin{figure}[!h]
    \begin{center}
        \includegraphics[width=0.8\textwidth]{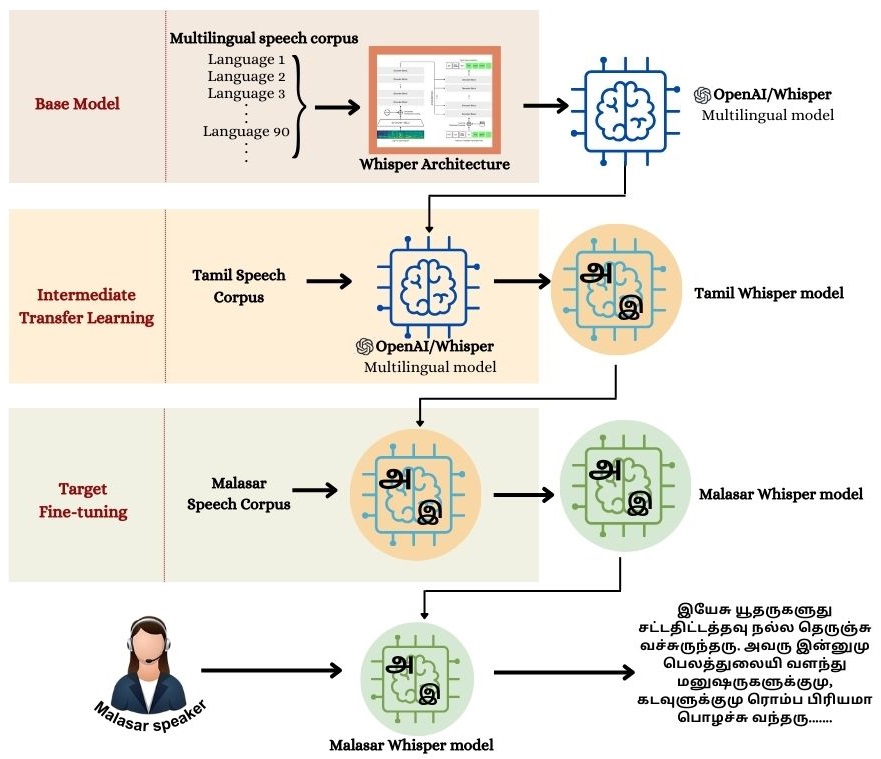}
        \caption{The proposed multistage fine-tuning strategy, where the Whisper architecture is first adapted to Tamil language and later to Malasar language.}
        \label{fig.fine}
    \end{center}
    \vspace{-0.35cm}
\end{figure}

To address the challenge of limited data in developing an ASR system for Malasar, we employed a novel multistage sequential fine-tuning approach using the Whisper model. This strategy, as illustrated in Fig. \ref{fig.fine}, was designed to leverage linguistic similarities between Malasar and Tamil, a language with more extensive resources and data availability.

In the first stage, we utilized an existing Whisper model already fine-tuned on Tamil as an intermediate model.  Both small and medium architectures of the Tamil fine-tuned models were used for this experiments. These intermediate models were fine-tuned on a diverse Tamil dataset that leverages several public ASR corpora, making it a suitable intermediate model due to its linguistic alignment with Malasar. The intermediate fine-tuning step allowed the model to better adapt to the specific characteristics of Malasar speech, despite the limited Malasar data available for training.

In the second stage, we fine-tuned this intermediate model further on the Malasar corpus, which comprises approximately four hours of audio transcribed in Tamil script. By introducing Malasar-specific features on top of the previously acquired Tamil linguistic structures, this multistage approach enabled more efficient model adaptation with limited data. 

\section{Result and Discussion}
In this section, we present and analyze the performance outcomes of our Malasar ASR system, focusing on the effectiveness of different model configurations and training strategies. By examining WER across these configurations, we aim to identify the optimal approach for improving ASR accuracy in a low-resource language context. 

\subsection{OpenAI Zeroshot and Direct Target Fine-tuning}
The zeroshot configuration, where the models are evaluated without any language-specific fine-tuning, provides a baseline WER of 115.4\% for Whisper Small\footnote{\url{https://huggingface.co/openai/whisper-small}} and 99.3\% for Whisper Medium\footnote{\url{https://huggingface.co/openai/whisper-medium}}. These high WER values highlight the significant challenges in recognizing Malasar speech directly, due to limited language similarity in the general model training data. The charcter error rates (CER)  of 58.3\% and 56.4\% for Whisper Small and Medium respectively, further illustrate the initial difficulty in capturing phonetic details of Malasar (Table. \ref{table:baseVsfine}).

\begin{table}[htpb]
  \caption{Evaluation results of Malasar test speech on on Whisper Small and Whisper Medium  after Zeroshot (\textit{no fine-tuning}) and Direct Target Fine-tuning (DTF)}
  \label{table:baseVsfine}
\begin{tabular}{lcccc}
\toprule
   \textbf{Configuration}  & \multicolumn{2}{c}{\textbf{OpenAI (Zeroshot)}} & \multicolumn{2}{c}{\textbf{Direct Target Fine-tuning}} \\
   & WER (\%)& CER (\%) & WER (\%) & CER (\%) \\
\midrule
Whisper Small   & 103.5 &  36.7 & 58.3 & 15.9\\ 
Whisper Medium   & 99.3 & 33.7 & 56.4 &15.8 \\ 
\bottomrule
\end{tabular}
\end{table}

When direct target fine-tuning is applied, the WER and CER drop notably for both small \footnote{\url{https://huggingface.co/vrclc/Malasar_small_DTF}} and medium \footnote{\url{https://huggingface.co/vrclc/Malasar_Medium_DTF}} models. This reduction demonstrates the effectiveness of target-specific adaptation, which allows the models to learn Malasar-specific phonetic patterns and linguistic structures, thereby improving recognition accuracy substantially.

\subsection{Multistage Target Fine-tuning (MTF) with Intermediate Tamil Pre-training}

In this approach, we used Whisper Small and Whisper Medium models pre-trained on Tamil data to leverage the phonetic and lexical similarities between Tamil and Malasar, followed by further fine-tuning on Malasar data. This intermediate transfer learning configuration yielded notable improvements in recognition accuracy over the zeroshot and direct target fine-tuning approaches. For the Whisper Small model, the Word Error Rate (WER) reduced significantly from its initial zeroshot level to 97.1\%, with a CER of 30.3\% after Tamil pre-training\footnote{\url{https://huggingface.co/vasista22/whisper-tamil-small}}. In the Whisper Medium model, the intermediate Tamil pre-training\footnote{\url{https://huggingface.co/vasista22/whisper-tamil-medium}} stage brought the WER down to 94.8\% and CER to 28.2\% (Table. \ref{table:inter}). These CER values indicate the benefits of transfer learning, as the models begin to capture Malasar’s phonological and structural patterns more effectively than with the baseline model.

\begin{table}[htpb]
  \caption{Evaluation results of Malasar test speech on Whisper Small and Whisper Medium after Intermediate Tamil Pre-training (\textit{Tamil pretraining}) and Multistage Target Finetuing}
  \label{table:inter}
\begin{tabular}{lcccc}
\toprule
   \textbf{Configuration}  & \multicolumn{2}{c}{\textbf{Intermediate Tamil}} & \multicolumn{2}{c}{\textbf{Multistage Target Finetuing}} \\
   & WER (\%)& CER (\%) & WER (\%) & CER (\%) \\
\midrule
Whisper Small   & 97.1 & 30.3 & 54.5 & 13.5\\ 
Whisper Medium   & 94.8 & 28.2 & 51.9 & 12.6 \\ 
\bottomrule
\end{tabular}
\end{table}

Following the intermediate transfer learning, the models were further fine-tuned on the Malasar dataset in a multistage target fine-tuning process. This stage resulted in further improvements, with the Whisper Small model\footnote{\url{https://huggingface.co/vrclc/Malasar_small_MTF}} achieving a WER of 54.7\% and a reduced CER of 13.5\%, while the Whisper Medium model\footnote{\url{https://huggingface.co/vrclc/Malasar_small_MTF}} reached a WER of 51.9\% and CER of 12.6\%. This reduction in WER and CER illustrates that multistage fine-tuning with related language pre-training enables the model to handle Malasar’s unique phonetic patterns more accurately.

\begin{table}[!h]
    \centering
    \caption{Samples of Reference Transcription and the ASR Prediction demonstrating the WER and CER. We illustrate the impact of the punctuation filter on the evaluation metrics.}
    \label{tab:normalization}
    \begin{tabular}{c}
         \includegraphics[width=1\textwidth]{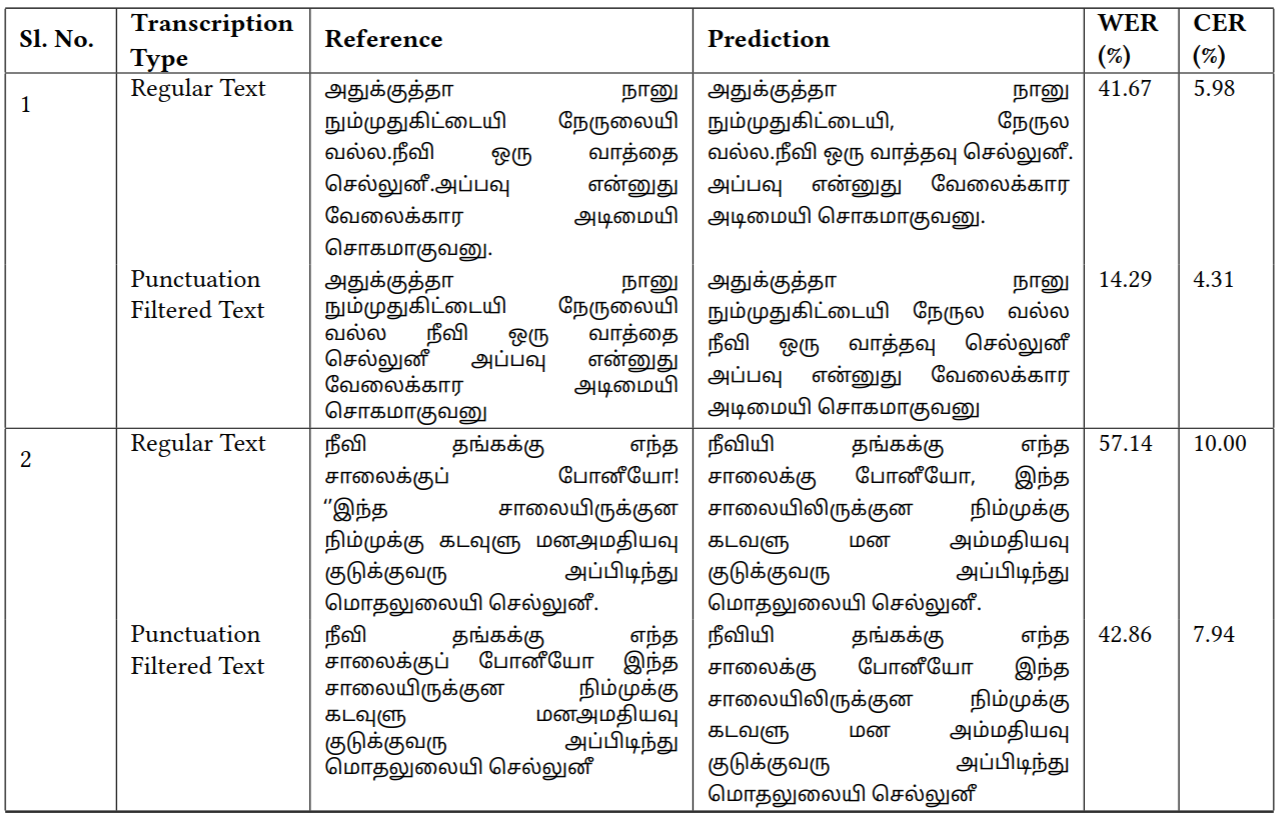}
    \end{tabular}

\end{table}
During model evaluation, it was observed that punctuation differences between the reference and predicted transcriptions significantly impacted WER and CER scores. To mitigate this, a normalization technique was introduced, which removes punctuation marks and symbols from both transcriptions before calculating error rates.

 Table \ref{tab:normalization} demonstrates sample transcriptions with and without punctuation filtering. For example, in the case of Sample 1, the WER decreased from 41.67\% to 14.29\% after filtering punctuation, and the CER dropped from 5.98\% to 4.31\%. Similar improvements were observed in Sample 2, where WER and CER values reduced from 57.14\% to 42.86\% and 10.00\% to 7.94\%, respectively, after normalization. These results underscore the importance of normalization in ASR evaluation for languages like Malasar, where inconsistent punctuation usage can inflate error metrics, potentially misleading ASR performance assessments. By removing punctuation, the evaluation becomes more focused on actual phonetic and lexical errors, offering a more accurate measure of the model's true effectiveness.

\begin{figure}[!h]
    \vspace{-0.35cm}
    \begin{center}
        \includegraphics[width=0.7\textwidth]{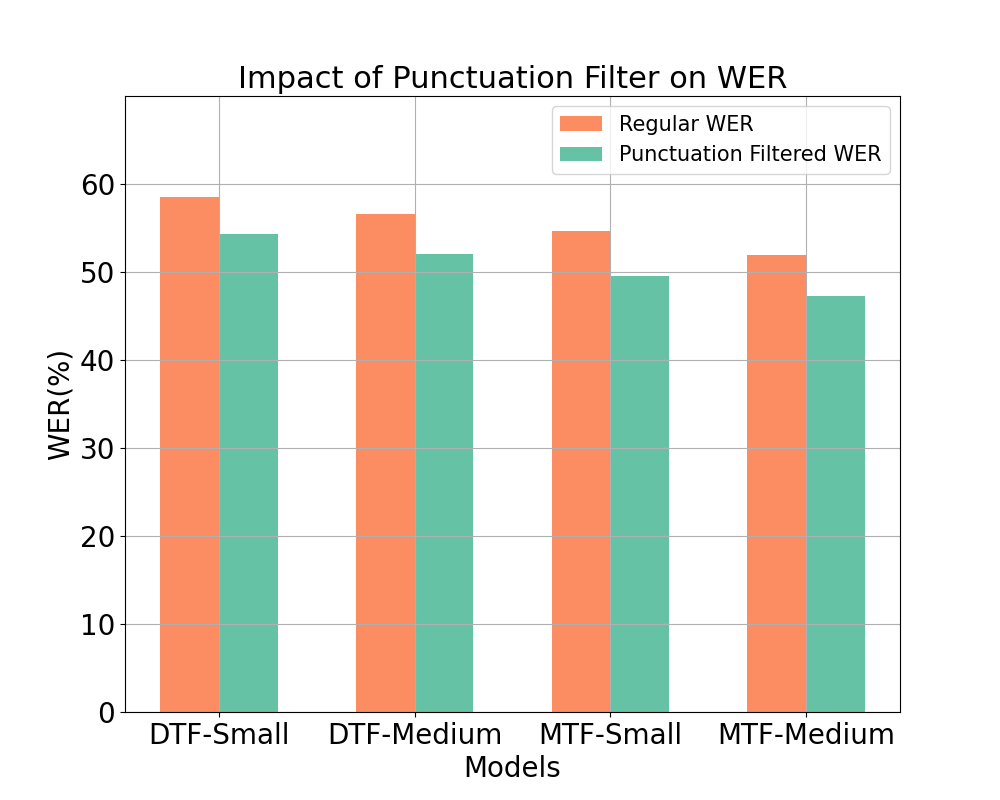}
        \caption{WER Evaluation on  direct target fine-tuning (DTF) and the proposed multistage target fine-tuning (MTF) before and after punctuation filtering.}
        \label{fig.WER}
    \end{center}
    
\end{figure}

This normalization applied during the evaluation of the Malasar ASR and the notable imporovements in performance is illustrated in Figure. \ref{fig.WER}. The initial results revealed a significant number of punctuation mismatches, particularly due to the nature of the corpus, which consists of biblical text containing numerous punctuation marks. We assessed the effectiveness of this two-stage fine-tuning approach by comparing it with direct fine-tuning on the Malasar data alone, with and without punctuation filtering, to demonstrate the contributions of each phase in achieving a lower WER. By introducing punctuation removal as a normalization technique, the WER for all models decreased notably. For instance, the fine-tuned Whisper-small model's WER improved from 58.5\% to 54.3\%, while the fine-tuned Whisper-medium model's WER dropped from 56.6\% to 52.1\%. The pretrained models with Tamil saw the most significant gains, with the fine-tuned pretrained Whisper-medium model achieving the best WER reduction from 51.9\% to 47.3\%. These results highlight the importance of normalization techniques, particularly in handling specialized corpora, and demonstrate that even simple normalizations like punctuation removal can significantly enhance ASR accuracy.

\section{conclusion}

This work presents a significant advancement in the development of ASR systems for low-resource languages through the innovative application of multistage fine-tuning techniques. By leveraging the linguistic similarities between the target language and a more resource-rich language, we have demonstrated that sequential fine-tuning can effectively enhance recognition accuracy and reduce word error rates. The results underscore the potential of using pretrained multilingual models as a foundation for ASR in languages with limited data, thereby contributing to the preservation and revitalization of endangered linguistic heritage. As this research lays the groundwork for future work, it opens avenues for further exploration into adaptive ASR strategies and highlights the crucial role of technology in empowering underrepresented languages, like Malasar, in the digital landscape. 

\bibliographystyle{ACM-Reference-Format}
\bibliography{sample-base}

\end{document}